\documentclass{article}

    \PassOptionsToPackage{numbers, compress}{natbib}

\usepackage[final]{neurips_2019}

\usepackage[utf8]{inputenc} 
\usepackage[T1]{fontenc}    
\usepackage{hyperref}       
\usepackage{url}            
\usepackage{booktabs}       
\usepackage{amsfonts}       
\usepackage{nicefrac}       
\usepackage{microtype}      

\usepackage{graphicx}
\usepackage{tabularx}
\usepackage{algorithm}
\usepackage{algorithmicx}
\usepackage{algpseudocode}
\usepackage{caption}

\usepackage{amssymb,amsmath,amsthm}
\usepackage{bm}

\DeclareMathOperator*{\argmin}{arg\,min}

\newcommand{\Ex}{\mathbb{E}}

\title{MetaCI: Meta-Learning for Causal Inference in a Heterogeneous Population}

\author{%
  Ankit Sharma \\
  TCS Research, Delhi\\
  \texttt{ankit.sharma16@tcs.com} \\
   \And
   Garima Gupta \\
   TCS Research, Delhi \\
   \texttt{gupta.garima1@tcs.com} \\
   \AND
   Ranjitha Prasad \\
   TCS Research, Delhi \\
   \texttt{ranjitha.prasad@tcs.com} \\
   \And
   Arnab Chatterjee \\
   TCS Research, Delhi \\
   \texttt{arnab.chatterjee4@tcs.com} \\
   \And
   Lovekesh Vig \\
   TCS Research, Delhi \\
   \texttt{lovekesh.vig@tcs.com} \\
   \And
   Gautam Shroff \\
   TCS Research, Delhi \\
   \texttt{gautam.shroff@tcs.com} \\}

\begin{document}

\maketitle

\begin{abstract}
Performing inference on data obtained through observational studies is becoming extremely relevant due to the widespread availability of data in fields such as healthcare, education, retail, etc. Furthermore, this data is accrued from multiple homogeneous subgroups of a  heterogeneous population, and hence, generalizing the inference mechanism over such data is essential. We propose the \texttt{MetaCI} framework with the goal of answering counterfactual questions in the context of causal inference (CI), where the factual observations are obtained from several homogeneous subgroups. While the CI network is designed to generalize from factual to counterfactual distribution in order to tackle covariate shift, \texttt{MetaCI} employs the meta-learning paradigm to tackle the shift in data distributions between training and test phase due to the presence of heterogeneity in the population, and due to drifts in the target distribution, also known as concept shift. We benchmark the performance of the \texttt{MetaCI} algorithm using the mean absolute percentage error over the average treatment effect as the metric, and demonstrate that meta initialization has significant gains compared to randomly initialized networks, and other methods.  
\end{abstract}

\section{Introduction}
Learning causal relationships is the heart and soul of several domains such as healthcare, advertising, education, economics, etc. For instance, personalized and targeted treatment considering an individual's health indicators is crucial in  healthcare~\cite{hanash2011emerging, seibold2016model}, targeted advertising campaign is essential to achieve higher profit margin in channel attribution~\cite{sun2015causal, taddy2016nonparametric,bottou2013counterfactual}. Causal inference (CI) aims to infer unbiased causality effect of the treatment from observational data by factoring the impact of the confounding variables of patients/users. In the context of observational studies, confounding variables affect the treatment and outcome, and hence, disentangling the effect of these variables is the key to achieve treatment effectiveness. In this work, we tackle the fact that the study-population is heterogeneous, and hence, developing CI-based systems that generalize for new unseen subgroups in data is essential in order to provide better targeted interventions.   

Classical approaches in CI estimate the average treatment effects from observational data by accounting for the selection bias using propensity scores, hence creating unbiased estimators of the averaged treatment effect (ATE)~\cite{rosenbaum1983central}. More recently, deep neural network based CI approaches have been proposed with different mechanisms to handle the bias. These include a latent variable modeling using VAEs~\cite{louizos2017causal}, a GAN-based technique~\cite{yoon2018ganite}, a DNN-based Deep IV~\cite{hartford2017deep}. In~\cite{johansson2016learning,shalit2017estimating}, the authors propose to view the causal inference problem as a covariate shift problem, and propose algorithms that balance between the factual and the counterfactual population.

Often, observational data is scarce, and the study-population is  heterogeneous. Subgroup analysis is proposed in literature for coping with heterogeneity in the population~\cite{vanderweele2019selecting,wager2018estimation}, especially in the context of establishing effect of the treatment for each subgroups~\cite{seibold2016model}. Our goal is to design a deep neural network based causal inference model that is capable of adapting/generalizing to new subgroups in the input data that may not have been encountered during training. To achieve this goal, we use the novel `learning to learn' paradigm, also known as the \emph{meta-learning} framework. Unlike conventional deep neural networks that require large amounts of data for training, meta-learning or few-shot learning learns to learn from previous \emph{tasks}, by discovering the structure among tasks to enable fast learning of new tasks~\cite{vanschoren2018meta}. In this work, we employ the algorithmic framework for CI proposed in~\cite{johansson2016learning,shalit2017estimating}, since it is a flexible framework in the context of meta-learning. 


\textbf{Contributions}: We apply the meta optimization based technique known as \emph{Reptile} on a well-known causal inference model~\cite{johansson2016learning}. 
A crucial design challenge is to define \textit{tasks}, as in meta-learning context, appropriately for a given problem. Specifically, we define tasks based on features of the subgroups in such a way that tasks contain some commonality w.r.t to subgroups. In scenarios that have multiple substructures in the deep neural network model, we propose the `multi-Reptile', which employs different learning rates for the parameters of the substructures.

As in~\cite{johansson2016learning}, we assume that there is no hidden confounding. We demonstrate the results on two datasets -- (a) synthetic dataset in the advertisement domain~\cite{sun2015causal}, and (b) semi-synthetic dataset based on the IHDP dataset ~\cite{hahn2019atlantic}. We employ mean absolute percentage error (MAPE) defined on ATE as the metric, and demonstrate that our \texttt{MetaCI} framework counters the effect of heterogeneity in the input population and handles the change in target distributions during inference time, while the CI network counters the issue of covariate shift.

\begin{figure}[t]
    \centering
     \hspace{-0.4cm}
    \includegraphics[width=5.3cm]{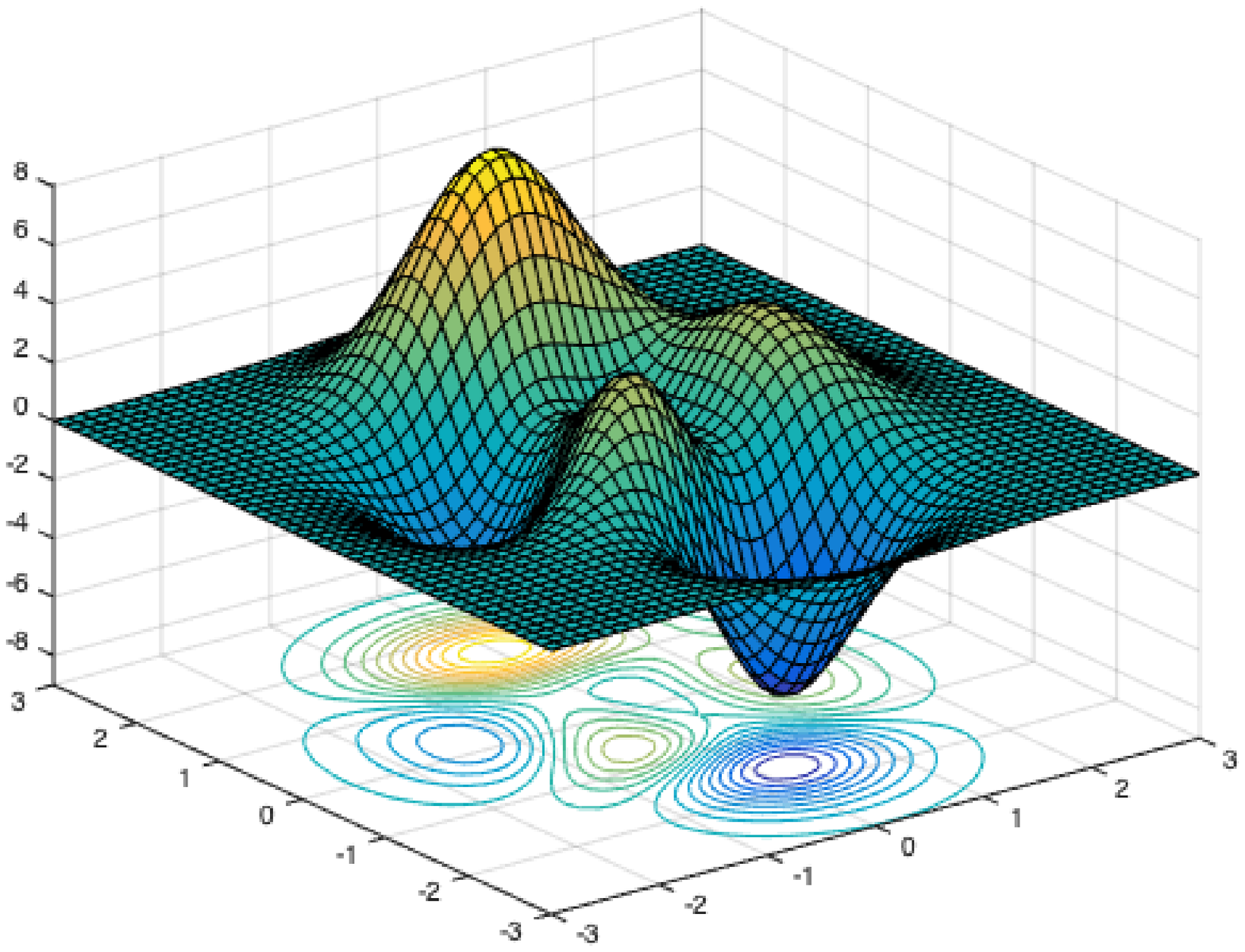} 
    \hspace{-0.4cm}
    \includegraphics[width=5.4cm]{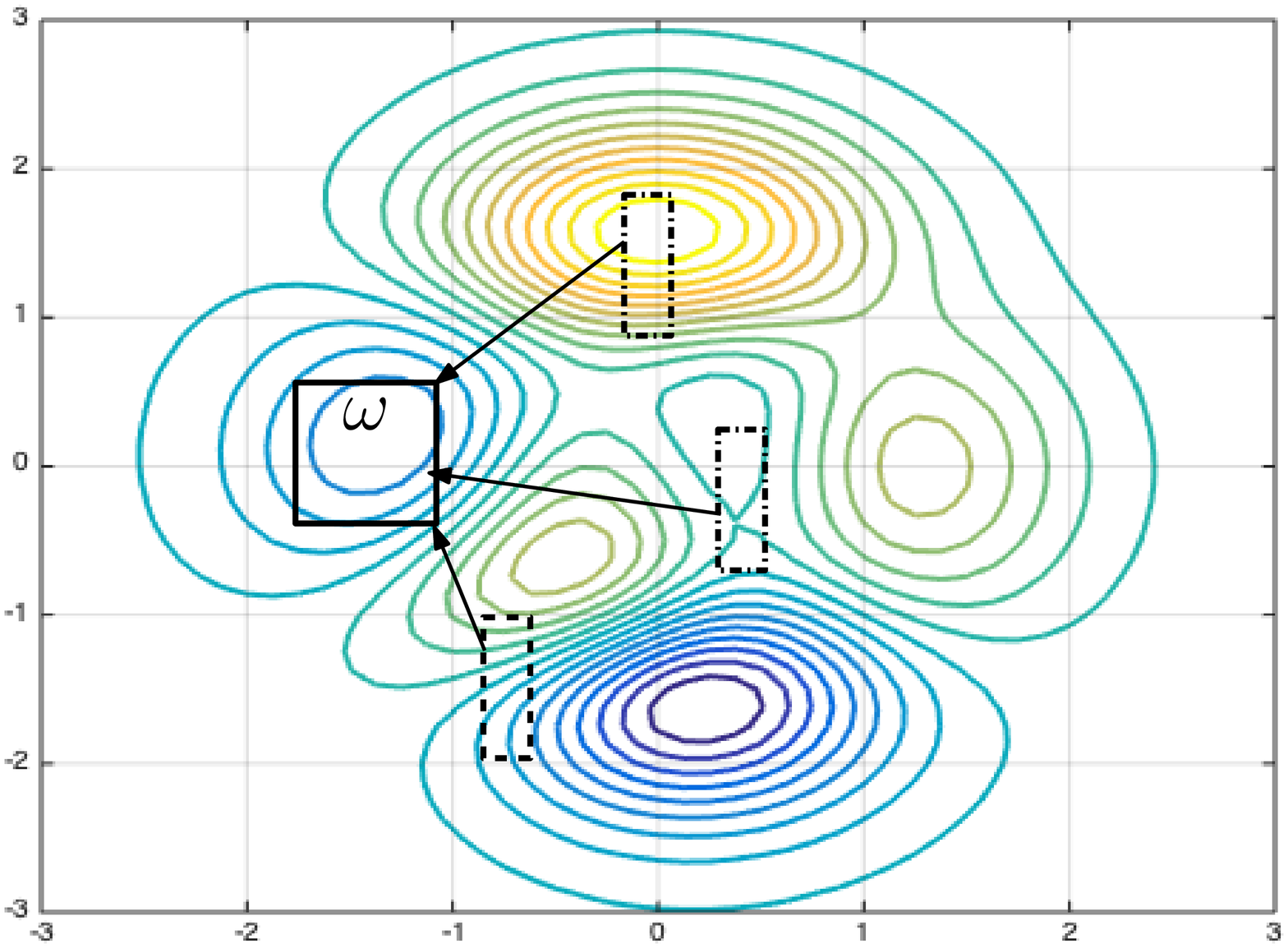}
    \hspace{-0.2cm}
    \includegraphics[width=3.3cm]{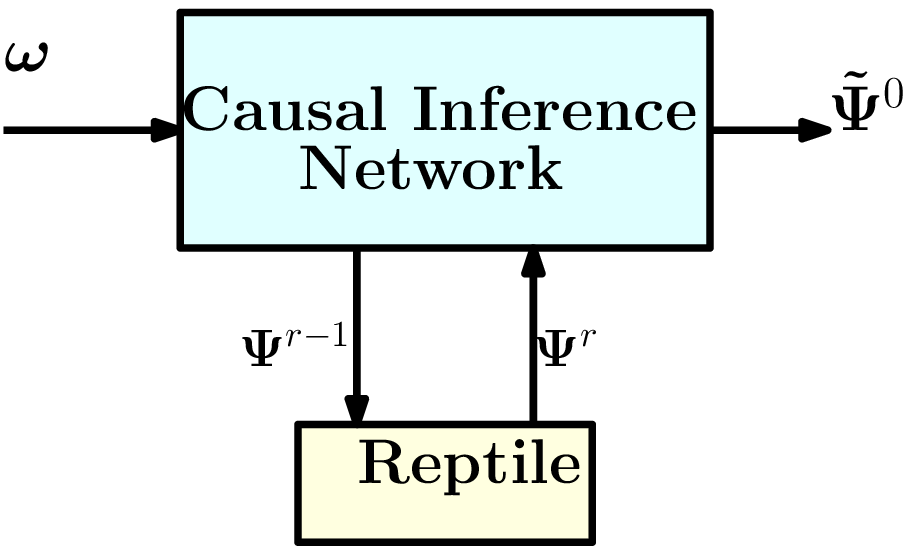}
    \caption{Toy example: For the joint distribution of the confounding variables (Left), task $\omega$ consists of samples in $X$ belonging to the region in the joint pdf (Centre). This region is contaminated by smaller non-overlapping regions of the joint pdf, in order to bring in commonality among tasks. Further, these tasks are the input to the \texttt{MetaCI} framework, to obtain the meta initialization given as $\bm{\tilde{\Psi}}^0$, as depicted above  (Right).}%
    \label{fig:toyexample}%
\end{figure}

\section{Preliminaries}
In this section, we describe \emph{Reptile}, an optimization based meta-learning paradigm, followed by description of the CI framework proposed in~\cite{johansson2016learning}.

\subsection{Meta-optimization preliminaries: Reptile}

Reptile is an optimization based approach to meta-learning, where the model parameters are adapted for fast learning with a few examples. In~\cite{nichol2018first}, the authors state the optimization problem in this context for an initial set of parameters $\Psi$, a randomly sampled task $\omega$ with corresponding loss given by $\mathcal{L}_{\omega}$, as follows 
\begin{align}
    \hat{\Psi} = \argmin_{\Psi} \Ex_{\omega}\left[\mathcal{L}_\omega(U_\omega^L(\Psi))\right],
    \label{eq:reptileoptimprob}
\end{align}
where $U^L_{\omega}(\cdot)$ is an update operator, and $L$ represents the stochastic gradient descent (SGD) epochs. 

As an algorithm, Reptile involves repeatedly sampling task $\omega$, followed by learning the parameters using an update operator (e.g., SGD) on the data pertaining to $\omega$, and updating these parameters by learning on different tasks. The training phase of this framework provides a meta-initialization for the parameters $\Psi$ of the network, such that, for a new unseen task, network can be fine-tuned using this meta-initialization and a small amount of data from a new task. We employ the parallel version of reptile, where the solution for the optimization problem in \eqref{eq:reptileoptimprob} is given by
\begin{equation}
    \Psi \leftarrow \Psi + \epsilon (\tilde{\Psi} - \Psi),
    \label{eq:reptile}
\end{equation}
where $\epsilon$ is an adaptive learning rate, and $\tilde{\Psi}$ is obtained after applying the update operator on the $\omega$-th task data. In this work, we consider the tasks pertaining to the causal inference where the goal is to learn a model for counterfactual inference. Hence, $U_\omega^L(\Psi)$ is a stochastic gradient descent operator which optimizes a cost function pertaining to counterfactual inference as given in~\cite{johansson2016learning}. We use the meta optimization framework to tackle both, the prior shift that occurs due to a drift in the feature distribution across tasks, and the concept shift that occurs due to a drift in probability distribution of the target variables~\cite{kouw2018introduction}. In the sequel, we provide the basic setting of a causal inference problem, and describe the CI network which we use as the update operator, $U_\omega^L(\Psi)$.

\subsection{Causal Inference preliminaries}

In this subsection, we describe the problem of counterfactual inference in the meta-optimization framework. The CI network that we employ was proposed in~\cite{johansson2016learning,shalit2017estimating}.

Let $\mathcal{T}$ represent the set of treatments, $\mathcal{X}^\omega$ be the set of contexts, and $\mathcal{Y}^\omega$ be the set of possible outcomes w.r.t. the $\omega$-th task. We assume that the treatment is binary, that is $\mathcal{T} \in \{0,1\}$, where we assign treatment $t = 1$ as \emph{treated} and  $t = 0$ as \emph{control}. Note that, for a given context $x^\omega \in \mathcal{X}^\omega$, we observe one of the potential outcomes $y^\omega \in \mathcal{Y}^\omega$, according to the treatment provided, i.e., if $t^\omega = 0$, we observe $y^\omega = Y^\omega_0$, and  if $t^\omega = 1$, we observe $y^\omega = Y^\omega_1$, and accordingly we are interested in optimizing the ITE for the context in task $\omega$, $x^\omega$ is given by $ITE(x^\omega) = Y^\omega_1(x^\omega) - Y^\omega_0(x^\omega)$. Furthermore, we are also interested in the the average treatment effect (ATE) averaged over all tasks and contexts, defined as $ATE = \Ex_{\omega \sim p(\omega)}\left[\Ex_{x^\omega \sim p(x^\omega)} ITE(x^\omega)\right]$. 


In~\cite{johansson2016learning}, the authors perform counterfactual inference by generalizing from the factual to counterfactual distribution. To this end, they learn a representation $\Phi^\omega$ and the function $h^\omega$, such that one term optimizes the prediction error w.r.t. the observed outcomes over the factual representation, the other term ensures that the distributions of treatment populations are similar or balanced for a given task $\omega$, , thus tackling the issue of covariate shift~\cite{shalit2017estimating}. Accordingly, the objective to minimize is
\begin{align}
    \mathcal{L}\left(\alpha^\omega, \gamma\right) &= \frac{1}{N_\omega} \sum_{i=1}^{N_\omega} w_i^\omega \mbox{L} {\left(h^\omega(\Phi^\omega(x^\omega_i),t_i^\omega) , y_i^{\textnormal{F},\omega}\right)} + \alpha^\omega \mbox{disc}\left(\hat{P}^{\textnormal{F}}_{\Phi^\omega},\hat{P}^{\textnormal{CF}}_{\Phi^\omega}\right) + \gamma \mathcal{R}(h^\omega),
\end{align}
where $\alpha^\omega, \gamma > 0$ are hyper-parameters that control the strength of the imbalance penalties, $w_i$ compensate for the difference in treatment group size,  $\mathcal{R}(h^\omega)$ is a model complexity term, $\hat{P}^{\textnormal{F}}_{(\cdot)}$ represents the factual distribution, and  $\hat{P}^{\textnormal{CF}}_{(\cdot)}$ represents the counterfactual distribution, respectively, and $\mbox{disc}(\cdot,\cdot)$ is the discrepancy measure as defined in~\cite{shalit2017estimating}.

\section{\texttt{MetaCI} Model}
In this section, we present the process of task creation, and describe the proposed \texttt{MetaCI} model.  

\subsection{Task creation}
\label{sec:taskCreation}
It is well known that a good meta-learning model should be trained for a diverse set of learning tasks and optimized based on the probability distribution of different tasks, including potentially unseen tasks. Defining task similarity is the key overarching challenge in meta learning. In the presence of heterogeneity in the population, we employ our knowledge regarding the features specific to subgroups, which are also the confounding variables in order to define tasks. We create tasks by combining a majority of samples from one subgroup, and a few samples from other subgroups in fixed proportions. Mathematically, using the joint distribution of the confounding variables, we ensure that we choose a subgroup that lies in a given region of the joint distribution, and mix it with samples from smaller disjoint regions of the same joint probability distribution, as depicted for a toy example in Fig.~\ref{fig:toyexample}. 

\section{Proposed Model}
\begin{figure}[t]
    \centering
    \includegraphics[width=7.0cm]{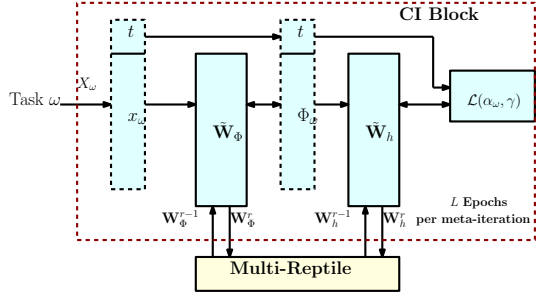}
    \caption{Block diagram describing the \texttt{MetaCI} framework for a given task $\omega$.}
 \label{fig:MetaCImodel}
\end{figure}

In this section, we propose a novel \texttt{MetaCI} algorithm, where we combine a variant of the Reptile framework along with the causal inference framework~\cite{johansson2016learning}. 
As depicted in the neural network model in Fig.~\ref{fig:MetaCImodel}, we see that sampling of task and the update of weights using Multi-Reptile meta-learning algorithm occurs outside the CI block. The CI block constitutes the update operator in the context of meta-learning framework, and $L$ SGD epochs are used per meta-iteration. We term the meta-learning variant as Multi-Reptile, since it employs multiple adaptive learning rates for different subset of parameters of the update operator $U^L_\omega(\mathbf{W})$. Specifically, in the case of the CI network, we employ different learning rate for the representation and the hypotheses layers. The \texttt{MetaCI} algorithm is formally stated in Algorithm.~\ref{alg:metaCI}.

\begin{algorithm}
\caption{MetaCI algorithm}
\label{alg:metaCI}
\begin{algorithmic}[1]
\Procedure{Meta-CI}{arguments}
\State {For all tasks, sample a test task $\omega \in \Omega_{te}$, and $\Omega_{tr}$ constitute the pool of train tasks.}
\For{$R$ iterations}
\State {Sample task $\omega \in \Omega_{tr}$.}
\State {Compute the weights $\tilde{\mathbf{W}}_{\Phi}$ and $\tilde{\mathbf{W}}_{h}$ using $U^L_{\omega}(\mathbf{W}_{\Phi})$ and $U^L_{\omega}(\mathbf{W}_{h})$, respectively.}
\State {Meta update weights of the representation layer:}
  $\mathbf{W}^{r+1}_{\Phi} = \mathbf{W}^r_{\Phi}+\epsilon_\Phi(\tilde{\mathbf{W}}_{\Phi} - \mathbf{W}^r_{\Phi}) $
\State {Meta update weights of the hypotheses layer:}
  $\mathbf{W}^{r+1}_{h} = \mathbf{W}^r_{h}+\epsilon_h(\tilde{\mathbf{W}}_{h} - \mathbf{W}^r_{h})$ 
\EndFor \\
\Return {$\mathbf{W}_{\Phi}$ and $\mathbf{W}_{h}$.}
\EndProcedure
\end{algorithmic}
\end{algorithm}


\section{Experiments}
In this section, we describe the datasets, the mechanism used for creating tasks for each dataset as described in  Sec.~\ref{sec:taskCreation}, followed by the metrics we employ for evaluation, and finally the experimental results.

\subsection{Datasets}
We demonstrate the performance of the proposed algorithms on a  synthetically generated advertisement dataset~\cite{sun2015causal} and the semi-synthetic IHDP dataset~\cite{hahn2019atlantic}, for evaluation \footnote{The simulated datasets will be available upon request from authors post publication of the paper.}.

\subsubsection{Synthetic advertisement dataset}
We use a synthetic data generating process (DGP) to generate data relevant to the advertisement domain, as described in~\cite{sun2015causal}. We set the sample size $N = 2000$ and number of features $p = 10$. We generate features $q_1, \hdots, q_P \sim \mathcal{N}(0,1)$, and the basis functions $f_1(x), \hdots, f_{10}(x)$ as described in~\cite{sun2015causal}. We restrict the  treatment $T$ as being binary, and generate the treatment as $T|X = 1$ if $ \sim \mathcal{N}(\sum_{j=1}^5 f_j(q_j),1) > 0$, and $0$ otherwise. Further, we generate the response as $Y|T,X \sim \mathcal{N}(\sum_{j = 1}^5 f_{j+5}(q_j)+\eta^T T,\theta)$. We set $\theta = 1$ to generate data for demonstrating the effect of covariate shift, and set $\theta$ as $1$, $10$ and $20$ to generate data for demonstrating the effect of concept shift. Note that the features $q_1, \hdots, q_5$ have  confounding effects on both the treatment and the outcome, and the rest of the features contribute to the noise in the model.

\subsubsection{Semi-synthetic IHDP dataset}
The Infant Health Development Program (IHDP)~\cite{IHDP} dataset consists of measurements of mother and children for studying the effect of specialist home visits on future cognitive test scores. The dataset comprises of $4302$ infants having $25$ features. Out of these, $8$ are selected based on ACIC challenge (2017) to obtain context information $X$. Specifically, these features form the basis of the meta-learning tasks obtained using the DGP~\cite{hahn2019atlantic}. 

\subsection{Task creation for Reptile}
Here we describe the process of task creation to demonstrate the performance of the \texttt{MetaCI} framework in the presence of covariate and concept shift, for the datasets provided in the previous section.

\subsubsection{Covariate shift}
\textbf{Tasks in synthetic dataset}: In order to appropriately provide tasks to the \texttt{MetaCI} framework in presence of covariate shift, we generate $2000$ users distinguished based on the set of features, for number of tasks defined by cardinality of $\Omega$. We consider these $|\Omega|$ disjoint chunks, and mix it with  samples from other chunks in the ratio $3:2$, i.e., each task consists of $60\%$ of samples from a given chunk, and $40\%$ of samples in equal proportion from $k$ other chunks. For every subgroup, $T|X$ and $Y|T,X$ is generated using a generating process specified in~\cite{sun2015causal}.
In the single feature case, the data is split on the basis of the first feature which is one of the confounding variables. In the case of multiple confounding features, the data is split on the basis of the first two features which are confounding. We create tasks based on the joint distribution of the confounding features as outlined in Sec.~\ref{sec:taskCreation}.

\textbf{Tasks in IHDP dataset}: Here we create tasks for the \texttt{MetaCI} framework for the IHDP dataset, with an end goal of demostrating the performance of the proposed algorithm in presence of covariate shift. We define tasks by dividing the entire population of infants, given as a finite number of contexts in the ACIC challenge dataset, 2017, into $|\Omega|$ equal sized chunks. We create these chunks based on the joint distribution of multiple confounding features. Specifically, we consider mother's age, child's bilirubin level and mother's place of birth. Each chunk is mixed with samples from other chunks in the ratio $3:2$, i.e., each task dataset, $X_{\omega}$, consists of $60\%$ of samples from a given chunk, and $40\%$ of samples in equal proportion from $k$ other chunks. For each of the tasks, $T$ and $Y_{\omega}$ is generated synthetically using hetroskedastic, additive error DGP given in~\cite{hahn2019atlantic}. 

In both the above cases, the number of chunks used for mixing ($k$) is an experimental variable and lies in range $[1,|\Omega|-1]$. 

\subsubsection{Concept and covariate shift}
\label{sec:tasksConceptShift}
\textbf{Tasks in synthetic and IHDP dataset scenario:} In order to demonstrate the performance of \texttt{MetaCI} in the presence of concept shift, we use two different generation processes 
which differ in generation of the response variable $Y$. Accordingly, we describe two types of task creation as follows:
\begin{enumerate}
    \item Case $1$- concept shift using $2$ DGPs: Based on the confounding features of the datasets, we consider $4$ chunks per DGP, and $3$ chunks per DGP, in synthetic and IHDP datasets, respectively.
    \item Case $2$- concept shift using $3$ DGPs: We consider $3$ chunks per DGP and $2$ chunks per DGP, in synthetic and IHDP datasets, respectively. 
\end{enumerate}

In both the above cases, the chunks are mixed within and across groups by retaining $60\%$ of the samples of one chunk, and replacing the remaining $40\%$ with samples from other chunks, to create tasks. The mixed chunks contribute to generating the responses as dictated by the number of DGPs. Across DGPs, the parameters of the distribution which is used to sample $Y|T,X$ is varied to demonstrate concept shift.

\subsection{Metrics}

In this subsection we describe the performance metrics used for evaluating proposed causal meta model. We use average treatment effect ($ATE_{\omega,r}$) for $r$-th test iteration and test task $\omega$ as the  performance metric, which is defined as
\begin{equation}
ATE_{\omega,r} = \frac{\sum_{i=1}^{N_{\omega,t_1}}(y_{i,1} - \hat{y}_{i,0})}{2N_{\omega,t_1}} + \frac{\sum_{i=1}^{N_{\omega,t_0}}(\hat{y}_{i,1} - y_{i,0})}{2N_{\omega,t_0}},
\end{equation}
where $y_{i,1}$ ($y_{i,0}$) is the factual response to treatment $t_i = 1$ ($t_i = 0$) and $\hat{y}_{i,0}$ ($\hat{y}_{i,1}$) is its corresponding counterfactual response, $N_{\omega,t_{1}}$ ($N_{\omega,t_{0}}$) are the number of samples in the task $\omega$ that are offered treatment $1$ ($0$). In order to eliminate any bias in the test set, we report the averaged $ATE$ corresponding to the iteration that has the least averaged validation objective across test set of the meta-test tasks. 
In the following section, we report the mean absolute percentage error defined on the ground truth ATE $ATE_G$, and the $ATE$ obtained as above as follows:
\begin{equation}
    MAPE = \left|\frac{ATE_G - ATE}{ATE_G}\right|,
\end{equation}
i.e., lower values of MAPE indicate that the obtained ATE values are closer to the ground truth ATE.

\subsection{Experimental details and results}
\begin{figure}[h]
    \centering
    \includegraphics[width=8.0cm]{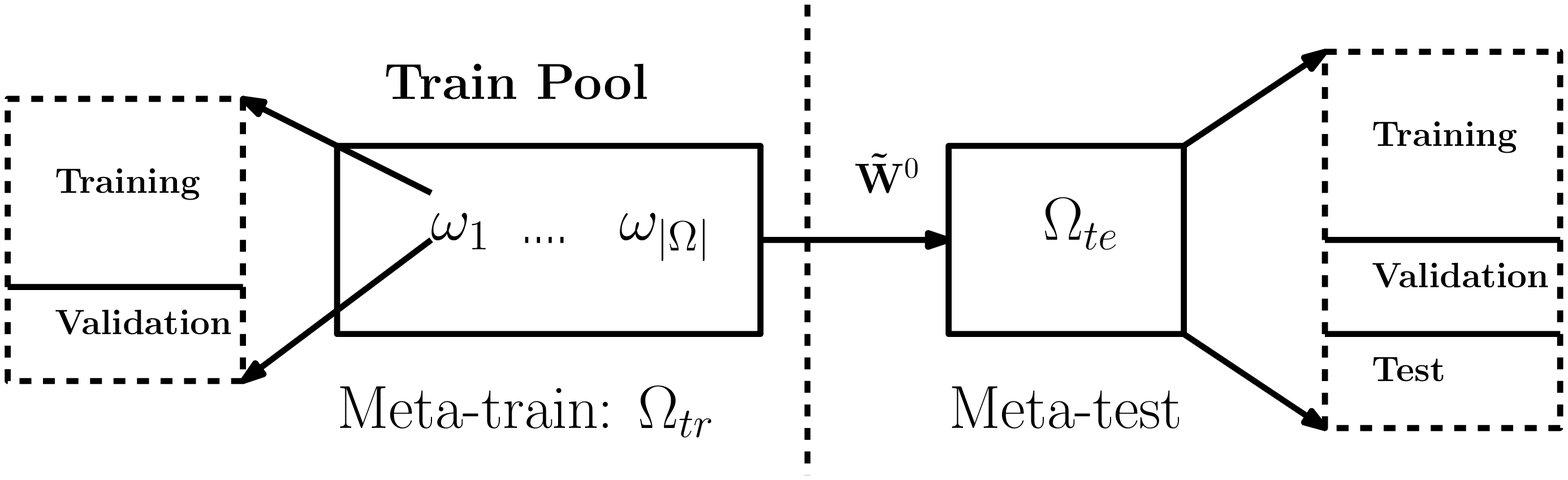}
    \caption{Block diagram describing the training procedure of \texttt{MetaCI}.}
 \label{fig:MetaCItrain}
\end{figure}

In this section, we report the experimental details and the results obtained.
We split $|\Omega|$ tasks into $|\Omega|-1$ train tasks and a test task as shown in Fig.~\ref{fig:MetaCItrain}. Every train task is divided in the ratio $1:1$ corresponding to training and validation and test task is divided in the ratio $2:1:1$ corresponding to training, validation and test sets. The \texttt{MetaCI} framework is trained for $1000$ iterations by sampling a train task in each iteration. For each iteration ($r$), weights ($W_{r}$) of causal meta model are computed after $L = 64$ epochs of mini-batch Stochastic Gradient Descent (SGD) over the batches of train set of train task.
These weights $W_{i}$ (where $r=i$ during training of \texttt{MetaCI}) are then used to update the initial weights $W_{0,i}$ present at the start of each iteration using reptile update Eq.~\eqref{eq:reptile}.

We pick the best train task hyper-parameters (learning rate, dropout, $\epsilon$) correspond to the least value of validation loss function averaged across all iterations. We evaluate the performance on the test set of test task (refer Fig.~\ref{fig:MetaCItrain}) by tuning the meta causal models' weights ($W_{0,j}$, where $j$ is every $100^\mathrm{th}$ iteration) for $64$ epochs on the test task's train set. Best hyper-parameters for test task is obtained in the same manner as discussed for training phase.

We repeat each experiment by considering each of $|\Omega|$ tasks as meta test tasks, and report the averaged MAPE across test sets of each test task as in Fig.~\ref{fig:MetaCItrain}.

We consider two baselines for \texttt{MetaCI}. The first baseline is meta learning based reptile algorithm that uses the \texttt{NN4} causal inference network. This baseline was presented in~\cite{johansson2016learning}. \texttt{NN4} does not incorporate a representation layer $\Phi$, as compared to the CI neural network in \cite{johansson2016learning}, and hence it is a good baseline . The authors demonstrate the superiority of their proposed network as compared to \texttt{NN4}. In the tables that follow in the next section, we employ two variants of this baseline, namely \texttt{MetaNN4}, which uses meta-initialization, and \texttt{RandomNN4}, which uses random initialization, both along with \texttt{NN4}. By adopting \texttt{NN4} along with meta learning, we verify that the gains obtained by using CI network as compared to \texttt{NN4} is carried over when we use meta learning.  In addition, we provide another baseline which consists of the CI network which is trained for large number of epochs over data from each task, but initialized using random initialization. This baseline helps us to gauge the performance of the CI network when the data is not provided in a meta learning fashion. We denote this baseline as \texttt{CI}$_\Omega$ in the tables that follow in the next section. 


\subsection{Results}

\begin{figure}
\centerline{\includegraphics[scale=0.7]{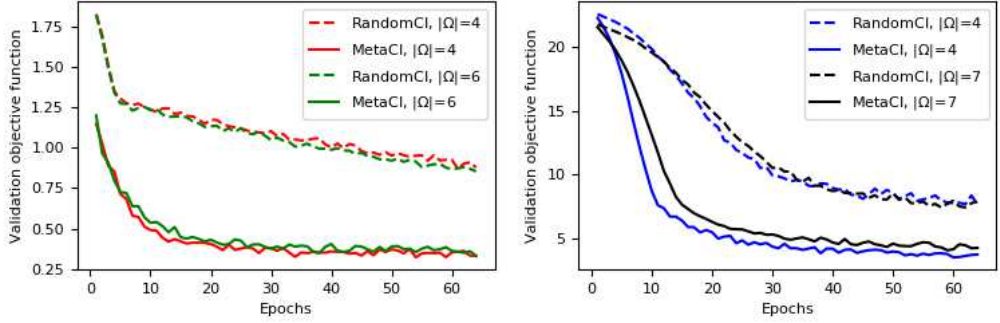}}
\caption{Comparison of validation objective (on test) across varying number of training epochs. (Left) IHDP dataset, (Right) Synthetic dataset.}\label{fig:Val_obj}
\end{figure}

We demonstrate the performance of \texttt{MetaCI} for varying number of tasks ($|\Omega|$), varying $k$, and $\epsilon$ using different settings for task creation, in the context of synthetic and semi-synthetic dataset discussed in previous section. We present the results pertaining to data that sees a covariate shift, and the combined effect of both, concept and covariate shift. Convergence is demonstrated in Fig.~\ref{fig:Val_obj}. 

\subsubsection{Covariate shift:}
\textbf{Varying number of subgroups ($|\Omega|$):} We study the performance by measuring the MAPE for varying number of tasks to study the effect of meta-initialization. In the context of synthetic dataset, we have the flexibility of generating as many samples as we require per task. Hence, in Table~\ref{Table_Varying_T_Single_Covariate} and \ref{Table_Varying_T_Multiple_Covariate} we set the number of samples per task to be same. However, the number of users are fixed in the case of the IHDP dataset, and hence, the number of samples per task goes down as the number of tasks increase. Furthermore, we set $k = |\Omega|-1$, i.e., as the number of tasks increase, the number of mixing chunks also increase, hence decreasing the commonality between tasks. Hence, we expect to observe a trade-off between data per task $N_{\omega}$ and $k$. From Table~\ref{Table_Varying_T_Single_Covariate} and 
Table~\ref{Table_Varying_T_Multiple_Covariate}, we see that this is indeed true, since we get the best MAPE for $|\Omega| = 7$ for single feature used for task creation in synthetic dataset case and $|\Omega| = 4$ , $|\Omega| = 6$ in case of multiple features used for task creation in IHDP and synthetic dataset respectively.
Furthermore, we see that the proposed technique performs better compared to the baselines described in the previous section. 

\begin{table}[h]
\caption{MAPE: Varying $|\Omega|$ ($k=|\Omega|-1$), using single feature for task creation. }
\label{Table_Varying_T_Single_Covariate}
\centering
     \begin{tabular}{cllllll}
      \hline
      \multirow{2}{*}{|$\Omega$|} 
              & \multicolumn{6}{c}{Synthetic dataset}\\\cline{2-7}
      & $N_\omega$ & $\texttt{CI}_{\Omega}$ & $\texttt{MetaCI}$ & $\texttt{MetaNN4}$ & $\texttt{RandomCI}$ & $\texttt{RandomNN4}$\\\hline
      4 &  2000 & 0.7305 & 0.3513 & 1.9400 & 1.5563 & 1.9991\\ \hline
      7 &  \textbf{2000} & \textbf{0.7088} & \textbf{0.2473} & \textbf{2.0993} & \textbf{1.3160} & \textbf{1.9348}\\ \hline
      9 &  2000 & 0.9487 & 0.4284 & 1.9995 & 1.2832 & 1.4622\\ \hline
      11 &  2000 & 0.8036 & 0.3475 & 0.7929 & 1.2855 & 0.9831\\ \hline
    \end{tabular}
\end{table}
\begin{table}[h]
\caption{MAPE: Varying $|\Omega|$ ($N_\omega$) using multiple features for task creation.}
\label{Table_Varying_T_Multiple_Covariate}
\centering
\begin{tabular}{cllll}
      \hline
      \multirow{2}{*}{|$\Omega$|} 
          & \multicolumn{3}{c}{IHDP}\\\cline{2-4}
      & $N_\omega$ & $\texttt{MetaCI}$ & $\texttt{RandomCI}$\\\hline
      4 & 1144 & 0.5164 & 1.6896\\ \hline
      6 & \textbf{764} & \textbf{0.5112} & \textbf{1.8492}\\ \hline
      8 & 498 & 1.7422 & 2.2367\\ \hline
    \end{tabular}
        \quad
\begin{tabular}{cllll}
      \hline
      \multirow{2}{*}{|$\Omega$|} 
          & \multicolumn{3}{c}{Synthetic dataset}\\\cline{2-4}
      & $N_\omega$ & $\texttt{MetaCI}$ & $\texttt{RandomCI}$\\\hline
      4 & \textbf{2000} & \textbf{0.3276} & \textbf{1.1786} \\ \hline
      7 & 2000 & 0.5528 & 0.6976\\ \hline
      9 & 2000 & 0.5762 & 0.6714\\ \hline
    \end{tabular}
\end{table}

\textbf{Varying number of chunks used for mixing ($k$):} We vary the number of mixing chunks $k$, for a fixed number of tasks $\Omega$, to study the effect of mixing on the performance as measured by MAPE. For $|\Omega| = 7$ and $|\Omega| = 6$, we see that varying $k$ leads to an improved value of ATE compared to the ground truth ATE in  
Table~\ref{Table_Varying_K_Multiple_Covariate}. 

\begin{table}
\caption{MAPE: Varying $k$ using single feature ($|\Omega|=7$) and multiple features ($|\Omega|=6$ and $|\Omega|=7$) for task creation. }
\label{Table_Varying_K_Multiple_Covariate}
\centering
\resizebox{\columnwidth}{!}{
    \begin{tabular}{clll}
      \hline
      \multirow{2}{*}{$k$} 
              & \multicolumn{2}{c}{Synthetic (single feature)}\\\cline{2-3}
      & $\texttt{MetaCI}$ & $\texttt{RandomCI}$ \\\hline
      3 & 0.4030 & 1.6603\\ \hline
      4 & 0.2490 & 1.4428\\ \hline
      6 & \textbf{0.2473} & \textbf{1.3160}\\ \hline
    \end{tabular}
\begin{tabular}{clll}
      \hline
      \multirow{2}{*}{$k$} 
          & \multicolumn{2}{c}{IHDP (multiple features)}\\\cline{2-3}
      & $\texttt{MetaCI}$ & $\texttt{RandomCI}$\\\hline
      2 & 0.7537 & 1.1847\\ \hline
      4 & \textbf{0.4546} & \textbf{1.7456} \\ \hline
      5 & 0.5112 & 1.8492  \\ \hline
    \end{tabular}
    \begin{tabular}{clll}
      \hline
      \multirow{2}{*}{$k$} 
          & \multicolumn{2}{c}{Synthetic (multiple features)}\\\cline{2-3}
      & $\texttt{MetaCI}$ & $\texttt{RandomCI}$ \\\hline
      3 & 0.8658 & 0.9750 \\ \hline
      4 & 0.8596 & 0.9263 \\ \hline
      6 & \textbf{0.5528} & \textbf{0.6976}\\ \hline
    \end{tabular}
  }
\end{table}

\textbf{Varying meta learning rate $\epsilon$:} We demonstrate the relative performance of multi-reptile, where we vary the relative weights ($\epsilon$) assigned to the parameters of the representation layer ($\mathbf{W}_\Phi$) vis-\'{a}-vis the weights assigned to the parameters of the hypotheses layer ($\mathbf{W}_h$). Across several scenarios and datasets, as shown in Table~\ref{Epsilon}, we observe that adopting a slower learning rate for the representation layer as compared to the hypotheses layer leads to $ATE$ very close to the ground truth ATE. Intuitively, the representation layer minimizes the discrepancy between distributions, which may vary slowly across tasks.

\begin{table}[!]
  \caption{Performance of the \texttt{MetaCI} framework for three scenarios, where speeds of relative weight adaptation of representation and hypotheses layer are varied.}
  \label{Epsilon}
  \centering
  \begin{tabular}{lllll}
  \hline
  Scenario: $|\Omega|=4, k=3$& $\epsilon_{h} \textgreater \epsilon_{\phi}$ & $\epsilon_{h} = \epsilon_{\phi}$ & $\epsilon_{h} \textless \epsilon_{\phi}$\\
  \hline
  Multiple co-variate IHDP dataset & \textbf{0.4994} & 0.5164 &  0.7169\\\hline
  Single co-variate synthetic dataset & \textbf{0.3230} & 0.3513 & 0.5840\\\hline
  Multiple co-variate synthetic dataset & 0.3621 & \textbf{0.3276} & 0.5966 \\\hline
  \end{tabular}
\end{table}

\subsubsection{Concept and covariate shift}

In this section, we present results for datasets in which we synthetically simulate concept and covariate shift at the same time. While covariate shift is inherent to the CI setting and arises due to confounding variables, concept shift arises due to the change in the probability distribution of the response variable conditioned on the input and treatment. 
In Table~\ref{table:ConceptShiftSynIHDP}, we demonstrate the performance of the \texttt{MetaCI} algorithm when there are $2$ and $3$ DGPs for generating the response as discussed in Sec.~\ref{sec:tasksConceptShift}. Mean ($\mu_{d}$) and variance ($\sigma_{d}^{2}$) of $ATEs$ per DGP for both the datasets shown in Table~\ref{table:ConceptShiftSynIHDP}, where $d$ = 1,2... We observed that \texttt{MetaCI} converges faster as compared to \texttt{RandomCI} for both the datasets. 

\begin{table}[!]
  \caption{Performance of \texttt{MetaCI} in case of covariate and concept shift using Synthetic and IHDP datasets.}
  \label{table:ConceptShiftSynIHDP}
  \centering
  \resizebox{\columnwidth}{!}{
  \begin{tabular}{lccccll}
  \hline
   & \# DGPs & ($\mu_{1}, \sigma^{2}_{1}$) & ($\mu_{2}, \sigma^{2}_{2}$) & ($\mu_{3}, \sigma^{2}_{3}$) & \texttt{MetaCI} & \texttt{RandomCI}\\
  \hline
 Synthetic & 2 & (0.4822, 0.0003) & (0.5356, 0.1739) & - & \textbf{0.4559} & 1.2523 \\\cline{2-7}
 & 3 & (0.5400, 0.0004) & (0.4733, 0.0337) & (0.7433, 0.4994) & \textbf{1.3153} & 2.0104 \\\hline
  \hline
 IHDP & 2 & (0.1515, 0.0002) & (0.9294, 0.0006) & - & \textbf{0.9419} & 1.3600 \\\cline{2-7}
  & 3 & (0.1521, 0.0007) & (0.9000, 0.0006) & (1.0288, 0.0080) & \textbf{1.6135} & 1.8699 \\\hline
  \end{tabular}
  }
\end{table}



\vspace{-2mm}
\section{Conclusions}
In this work, we demonstrate the efficacy of the meta learning based reptile framework in a causal inference setting for a heterogeneous population. We showed that the meta learning approach is a modern approach that could replace the classical subgroup analysis, where these subgroups can be translated as tasks in the meta learning framework. We provided a novel approach to create tasks based on the confounding features, and showed that it is possible to obtain a good meta initialisation which leads to significant improvement in ATE on the unseen data. We also showed that the \texttt{MetaCI} framework adapts its parameters in the presence of both covariate and concept shift in the dataset, and outperforms the baselines by large margins. We allude to specific details regarding training meta learning based deep neural network models, which by itself is a contribution to current literature.


\begin{thebibliography}{99}

\providecommand{\natexlab}[1]{#1}
\providecommand{\url}[1]{\texttt{#1}}
\expandafter\ifx\csname urlstyle\endcsname\relax
  \providecommand{\doi}[1]{doi: #1}\else
  \providecommand{\doi}{doi: \begingroup \urlstyle{rm}\Url}\fi
  
\bibitem{hanash2011emerging}
Samir~M Hanash, Christina~S Baik, and Olli Kallioniemi.
\newblock Emerging molecular biomarkers—blood-based strategies to detect and
  monitor cancer.
\newblock {\em Nature reviews Clinical oncology}, 8(3):142, 2011.

\bibitem{seibold2016model}
Heidi Seibold, Achim Zeileis, and Torsten Hothorn.
\newblock Model-based recursive partitioning for subgroup analyses.
\newblock {\em The international journal of biostatistics}, 12(1):45--63, 2016.

\bibitem{sun2015causal}
Wei Sun, Pengyuan Wang, Dawei Yin, Jian Yang, and Yi~Chang.
\newblock Causal inference via sparse additive models with application to
  online advertising.
\newblock In {\em Twenty-Ninth AAAI Conference on Artificial Intelligence},
  2015.

\bibitem{taddy2016nonparametric}
Matt Taddy, Matt Gardner, Liyun Chen, and David Draper.
\newblock A nonparametric bayesian analysis of heterogenous treatment effects
  in digital experimentation.
\newblock {\em Journal of Business \& Economic Statistics}, 34(4):661--672,
  2016.

\bibitem{bottou2013counterfactual}
L{\'e}on Bottou, Jonas Peters, Joaquin Qui{\~n}onero-Candela, Denis~X Charles,
  D~Max Chickering, Elon Portugaly, Dipankar Ray, Patrice Simard, and
  Ed~Snelson.
\newblock Counterfactual reasoning and learning systems: The example of
  computational advertising.
\newblock {\em The Journal of Machine Learning Research}, 14(1):3207--3260,
  2013.

\bibitem{rosenbaum1983central}
Paul~R Rosenbaum and Donald~B Rubin.
\newblock The central role of the propensity score in observational studies for
  causal effects.
\newblock {\em Biometrika}, 70(1):41--55, 1983.

\bibitem{louizos2017causal}
Christos Louizos, Uri Shalit, Joris~M Mooij, David Sontag, Richard Zemel, and
  Max Welling.
\newblock Causal effect inference with deep latent-variable models.
\newblock In {\em Advances in Neural Information Processing Systems}, pages
  6446--6456, 2017.

\bibitem{yoon2018ganite}
Jinsung Yoon, James Jordon, and Mihaela van~der Schaar.
\newblock Ganite: Estimation of individualized treatment effects using
  generative adversarial nets.
\newblock 2018.

\bibitem{hartford2017deep}
Jason Hartford, Greg Lewis, Kevin Leyton-Brown, and Matt Taddy.
\newblock Deep iv: A flexible approach for counterfactual prediction.
\newblock In {\em Proceedings of the 34th International Conference on Machine
  Learning-Volume 70}, pages 1414--1423. JMLR. org, 2017.

\bibitem{johansson2016learning}
Fredrik Johansson, Uri Shalit, and David Sontag.
\newblock Learning representations for counterfactual inference.
\newblock In {\em International conference on machine learning}, pages
  3020--3029, 2016.

\bibitem{shalit2017estimating}
Uri Shalit, Fredrik~D Johansson, and David Sontag.
\newblock Estimating individual treatment effect: generalization bounds and
  algorithms.
\newblock In {\em Proceedings of the 34th International Conference on Machine
  Learning-Volume 70}, pages 3076--3085. JMLR. org, 2017.

\bibitem{vanderweele2019selecting}
Tyler~J VanderWeele, Alex~R Luedtke, Mark~J van~der Laan, and Ronald~C Kessler.
\newblock Selecting optimal subgroups for treatment using many covariates.
\newblock {\em Epidemiology}, 30(3):334--341, 2019.

\bibitem{wager2018estimation}
Stefan Wager and Susan Athey.
\newblock Estimation and inference of heterogeneous treatment effects using
  random forests.
\newblock {\em Journal of the American Statistical Association},
  113(523):1228--1242, 2018.

\bibitem{vanschoren2018meta}
Joaquin Vanschoren.
\newblock Meta-learning: A survey.
\newblock {\em arXiv preprint arXiv:1810.03548}, 2018.

\bibitem{hahn2019atlantic}
P~Richard Hahn, Vincent Dorie, and Jared~S Murray.
\newblock Atlantic causal inference conference (acic) data analysis challenge
  2017.
\newblock {\em arXiv preprint arXiv:1905.09515}, 2019.

\bibitem{nichol2018first}
Alex Nichol, Joshua Achiam, and John Schulman.
\newblock On first-order meta-learning algorithms.
\newblock {\em arXiv preprint arXiv:1803.02999}, 2018.

\bibitem{kouw2018introduction}
Wouter~M Kouw.
\newblock An introduction to domain adaptation and transfer learning.
\newblock {\em arXiv preprint arXiv:1812.11806}, 2018.

\bibitem{IHDP}
Ruth~T. Gross.
\newblock Infant health and development program (ihdp): Enhancing the outcomes
  of low birth weight, premature infants in the united states, 1985-1988.
\newblock {\em MI: Inter-university Consortium for Political and Social
  Research}, 1993.

\end{thebibliography}
\end{document}